\documentclass[final,3p,times]{elsarticle}

\usepackage{amssymb}
\usepackage{amsmath}
\usepackage{amsfonts}
\usepackage{multirow}
\usepackage{booktabs}
\usepackage{times}
\usepackage{latexsym}
\usepackage{hyperref}
\usepackage{afterpage}

\usepackage{enumitem}
\usepackage{graphicx}
\usepackage{url}
\usepackage{hyperref}
\usepackage{longtable}
\usepackage{bm}

\usepackage{lineno}

\journal{arXiv}

\begin{document}

\begin{frontmatter}



\title{Deep Survival Analysis for Competing Risk Modeling with Functional Covariates and Missing Data Imputation}


\author[a]{Penglei Gao}
\affiliation[a]{
  organization={Department of Quantitative Health Science, Cleveland Clinic, Cleveland OH, USA},
}

\author[a]{Yan Zou}

\author[b]{Abhijit Duggal}
\affiliation[b]{
  organization={Department of Pulmonary and Critical Care Medicine, Cleveland Clinic, Cleveland OH, USA},
  }

\author[a]{Shuaiqi Huang}

\author[c]{Faming Liang}
\affiliation[c]{%
  organization={Department of Statistics, Purdue University, Lafayette IN, USA}
}

\author[a]{Xiaofeng Wang\corref{*}}
\cortext[*]{Corresponding author: Department of Quantitative Health Sciences, Cleveland Clinic, 9500 Euclid Ave/JJN3, Cleveland, OH 44195, USA. Email: wangx6@ccf.org}

\begin{abstract}
We introduce the Functional Competing Risk Net (FCRN), a unified deep-learning framework for discrete-time survival analysis under competing risks, which seamlessly integrates functional covariates and handles missing data within an end-to-end model. By combining a micro-network Basis Layer for functional data representation with a gradient-based imputation module, FCRN simultaneously learns to impute missing values and predict event-specific hazards. Evaluated on multiple simulated datasets and a real-world ICU case study using the MIMIC-IV and Cleveland Clinic datasets, FCRN demonstrates substantial improvements in prediction accuracy over random survival forests and traditional competing risks models. This approach advances prognostic modeling in critical care by more effectively capturing dynamic risk factors and static predictors while accommodating irregular and incomplete data.

\end{abstract}


\end{frontmatter}



\section{Introduction}\label{sec-intro}

Competing risks modeling is essential in time-to-event analysis when individuals are at risk of experiencing one of several mutually exclusive events, and the occurrence of one type of event prevents the occurrence of others~\citep{andersen2012competing,wolbers2014competing,austin2016introduction}. Traditional survival methods like Kaplan-Meier or Cox models may overestimate the probability of the event of interest when competing risks are present. Instead, competing risks models, such as the cumulative incidence function (CIF) and Fine-Gray sub-distribution hazard model~\citep{fine1999proportional}, offer more accurate tools for accounting for these events. 

Significant advancements have been made in discrete-time survival models to address competing risks in
the past decades \citep{schmid2021competing,berger2020subdistribution,lee2018analysis}. Unlike continuous-time approaches, such as the Fine-Gray sub-distribution model and the cause-specific hazard model~\citep{lau2009competing}, discrete-time competing risks models partition follow-up into fixed intervals (weeks, months, etc.) and analyze, for each interval, whether a particular event occurs first~\citep{heyard2019dynamic}. Right-censoring is handled naturally: once a subject is lost to follow-up or reaches the administrative end of the study, all subsequent person-intervals are simply omitted from the likelihood. This interval-based representation enables us to fit competing-risk models with standard generalized linear model machinery while retaining full flexibility for time-varying covariates and non-proportional effects.

Our motivating application is the real-time prediction of intensive-care-unit (ICU) readmission and post-discharge mortality using large-scale electronic health records (EHRs).  Numerous studies demonstrate that patients who return to the ICU have markedly longer hospitalisations and higher mortality rates \citep{amagai2025epidemiology,kumar2024determinants,ponzoni2017readmission,mcneill2020impact,lin2024prognostic}. Reported readmission frequencies range from 0.9 \% to 19 \%, and in-hospital mortality among these patients spans 13.3 \%–58 \% \citep{kumar2024determinants}.  Because of this clinical and economic burden, readmission has become a key quality indicator, and accurate prognostic tools are essential for safe discharge planning.

ICU time-to-event prediction presents several challenges due to data complexity and heterogeneity, including

\begin{itemize}
    \item {Functional covariates}: Continuous bedside monitors generate high-frequency time series (e.g., arterial pressure, heart rate), which can be viewed as functional data. These dynamic signals carry rich clinical information but are difficult to incorporate into conventional prediction modeling frameworks.
    \item {Missing data}: Laboratory values, medication doses, and clinical demographics are often missing or irregularly recorded; ignoring these missing values can introduce serious bias.
    \item {Data integration}: ICU datasets combine structured demographics, static clinical attributes, and multi-resolution time series. A successful model needs to harmonize these disparate modalities within a single predictive framework.
\end{itemize}

To tackle these issues, we introduce a novel deep learning framework, termed the Functional Competing Risk Net (FCRN), designed to address the challenges inherent in real-world medical data, particularly those involving functional covariates and missing data. The core architecture of FCRN consists of deep neural networks equipped with Fully Connected (FC) layers and non-linear activation functions. These components allow the model to capture intricate relationships within the data and effectively model complex patterns associated with competing risks. For the output layer, we consider two distinct hazard functions as output activation functions, enabling the estimation of both cause-specific and sub-distribution models. To handle the challenges associated with functional covariates, we introduce a novel preprocessing module that efficiently encodes the time-varying nature of these covariates. Specifically, we utilize a micro-network that embeds the time points of the functional data into a meaningful representation~\citep{yao2021deep}. The Basis Layer (BL) in this network transforms the functional covariates into embedding vectors, which capture the temporal structure and individual characteristics of the time series. These embedding vectors are integrated with the clinical variables and are processed by the main network to generate the final predictions. Our approach allows the model to handle heterogeneous data types simultaneously, enhancing its capacity to model real-world medical datasets where both time-dependent and static features coexist.
Moreover, the presence of missing data is a common issue in medical datasets and requires appropriate handling for ensuring the reliability and performance of predictive models. To address this challenge, we propose a second preprocessing module known as Missing Value Imputation (MVI). The MVI module leverages the Imputation-regularized Optimization (IRO) algorithm, which integrates missing data imputation directly into the network's training process~\citep{liang2018imputation}. By performing imputation and prediction simultaneously, the IRO algorithm ensures that the model accounts for uncertainty due to missing data while optimizing the predictive accuracy of the network. This joint process not only improves the imputation of missing values but also enhances the overall robustness of the model, helping ensure that the predictions are reliable even in the presence of incomplete data. Our contribution can be summarized as follows:
\begin{itemize}
    \item We propose FCRN, a unified deep-learning framework for competing risks modeling, showing promising performance on simulated datasets and two real-world ICU datasets. To our knowledge, FCRN is the first such framework to handle tabular and functional data while processing missing values end-to-end.
    \item We incorporate a Basis Layer with a micro neural network, learned jointly with the main network, to process functional covariates, effectively capturing trends and fluctuations for improved predictions.
    \item We integrate a gradient-based imputation algorithm with I-Step (Imputation) and RO-Step (Parameter Update) to impute missing values based on observed data and network outcomes.
\end{itemize}

 In our real application, we evaluate FCRN on two large‐scale ICU cohorts: (i) MIMIC-IV (Training and Testing cohort) is a publicly available database that contains de-identified EHRs for more than 65,000 critical-care admissions to Beth Israel Deaconess Medical Center between 2008 and 2019, including high-resolution bedside waveforms, laboratory results, interventions, and outcome labels~\citep{johnson2023mimic}. (ii) Cleveland Clinic ICU dataset (Validation cohort) comprises all adult patients admitted to any ICU in the Cleveland Clinic Health System during 2023; the dataset combines structured EHR fields with continuous vital-sign streams collected at the bedside.

 The remainder of the article is organized as follows. Section 2 reviews existing related work on deep learning for survival analysis. Section 3 presents the proposed FCRN architecture, including the Basis Layer for functional covariates and the gradient-based missing-value imputation scheme. Section 4 describes the experimental setup, datasets, and evaluation metrics, followed by results on synthetic and real ICU data. Section 5 discusses practical implications and limitations and concludes the paper. All software code files in this study are available at the GitHub through the link: \url{https://github.com/ClinicalAIML/Competing-Risk-Deep-learning-FCRN}.

\section{Literature Review}

In ICU settings, waveform measurements are often collected as time series, which can be treated as functional data. Functional Data Analysis (FDA) provides a principled framework for analyzing such data as functions \citep{ramsay2005functional, wang2016functional, ren2023multivariate}. A common approach in FDA is to represent functional observations using basis expansions. Let $\{\phi_d(\tau)\}_{d=1}^D$ be a set of $D$ continuous basis functions defined on the interval $[0,1]$. The functional signal $\tilde{x}(\tau)$ is then approximated as

\begin{equation}
\label{eq:fda}
\tilde{x}(\tau) \approx \sum_{d=1}^D a_d \phi_d(\tau),    
\end{equation}
where the coefficient vector $\bm{\mathrm{v}}_a = [a_1, \ldots, a_d]^\top$ summarizes the function in a finite-dimensional space. Common choices of basis functions include Fourier bases, B-splines, or eigenfunctions obtained from the spectral decomposition of the covariance function $\text{Cov}(\tilde{x}(\tau'), \tilde{x}(\tau))$.

Extending this idea, \citet{rossi2005functional} proposed using basis representations with multi-layer perceptrons (MLPs), treating the finite-dimensional vector $\bm{\mathrm{v}}_a$ instead of $\tilde{x}(\tau)$ as inputs to neural networks. More recently, \citet{yao2021deep} introduced a Basis Layer architecture in which each hidden unit is a micro neural network, effectively serving as a learnable basis function. These neural extensions allow for more flexible, data-driven representations of complex functional signals beyond traditional fixed basis sets.

Deep learning has increasingly addressed survival data beyond traditional statistical methods. \citet{biganzoli2006artificial} developed PLANNCR, a partial logistic artificial neural network for discrete survival times using multinomial likelihood, with time intervals as ordinal inputs. \citet{huang2021deepcompete} proposed DeepCompete, a continuous-time deep neural network (DNN) for competing risks that learns risks data-driven without strong stochastic assumptions. However, these methods do not simultaneously handle tabular and functional data with missing values in end-to-end training.

\section{Method}
\begin{figure}
    \centering
    \includegraphics[width=0.7\linewidth]{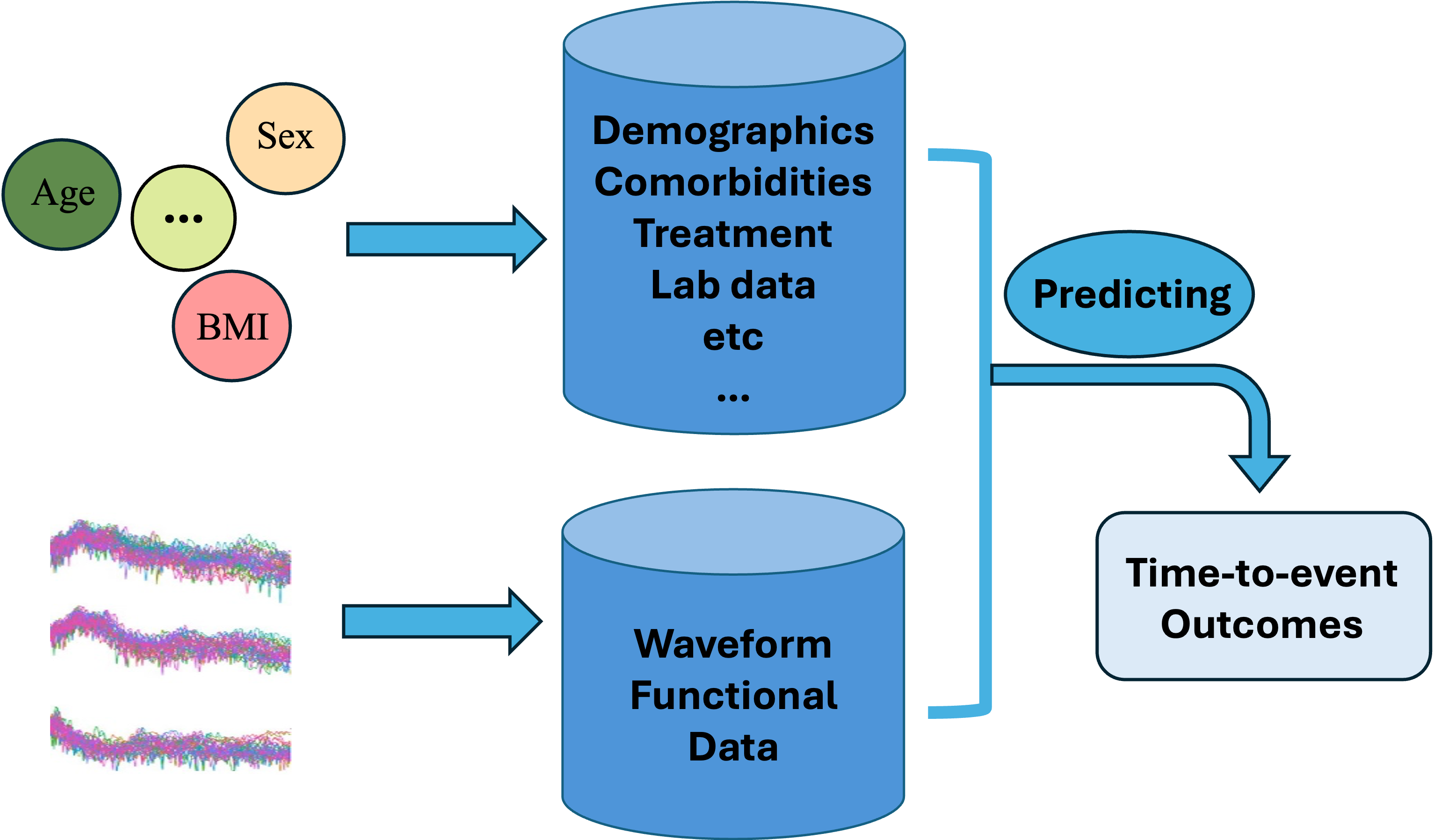}
    \caption{ICU data overview and the prediction flow.}
    \label{fig:data-overview}
\end{figure}

In this study, we aim to enhance our understanding of patient outcomes in ICU discharge by modeling two significant events: ICU readmission and death following discharge, which are important in decision-making and ICU prevention. First, it could enhance patient care by identifying individuals at high risk for adverse outcomes early, thus enabling timely and targeted interventions. Second, understanding these risks can help in optimizing ICU discharge planning and follow-up care, potentially reducing readmission rates and associated healthcare costs.
Our method is designed to effectively address the substantial challenges posed by complex ICU patient data, characterized by demographic predictors and the dynamic nature of patient conditions. In \autoref{fig:data-overview}, we present the general data types and flow chart for our ICU time-to-event prediction.
In the following, we will demonstrate the data preparation and detailed components of our method.

\subsection{Problem Statements}
Given a dataset containing $N$ patients, considering the usual right censoring situation, we define the observed event time as $T_i$ and the censoring time $C_i$ of a specific sample $i$, $i=1,2,...,N$. Let $R_i \in \{1,...,M\}$ denote the possible cause of an event that each subject $i$ is exposed to at risk of failure. In a time-to-event setting with competing risks, the observed event times $\Tilde{T}_i=min\{T_i, C_i\}$ are determined with a corresponding status indicator $\delta_i$, where $\delta_i=0$ indicates the censoring time and $\delta_i=1$ corresponds to the true event time,
\begin{equation*}
    \delta_i =
    \begin{cases}
        1& \quad T_i \leq C_i\\
        0& \quad T_i > C_i.
    \end{cases}
\end{equation*}
The observed cause of event $R_i$ is 0 if and only if $\delta_i=0$. The covariate vector $x_i$ comprises data on demographic, clinical, pathological, and biological characteristics of the patient, as well as information related to the administered treatment. The waveform functional data is represented by the time series vector $\tilde{x}_i$ for a given subject $i$. 

A key quantity for competing risks analysis is the discrete-time survival model. These statistical models are used to analyze time-to-event data where the time variable is measured in distinct, separate intervals rather than on a continuous scale. In these models, time is considered in discrete units such as weeks, months, or years. In our setting, the data record of subject $i$ is created for each time period from $t=1$ up to their event or censoring time $\Tilde{T}_i=min\{T_i, C_i\}$. Each record includes the time period $t$, the event indicator $\delta_i$, the observed cause of the event $R_i$, and the covariate pair $(x_i,\tilde{x}_i)$.
In traditional competing risks modeling, the cumulative incidence function (CIF) of each event $m \in \{1,...,M\}$ of subject $i$ is defined by $F_m(t|x_i):=P(T_i \leq t, R_i=m|x_i)$. The value of $F_m$ is between 0 and 1 by definition. CIF estimates the probability of a specific type of event $m$ occurring by a certain time $t$, in the presence of other competing events. It provides a way to describe the risk of an event happening over time while acknowledging that other types of events can preclude the occurrence of the event of interest.

Assume that $T_i$ and $C_i$ are independent random variables taking discrete values in $\{1,...,L\}$. In the discrete-time modeling, the originally continuous event times are grouped into $l=1,...,L$ disjoint intervals $(0,t_1], (t_1,t_2],..., (t_{L-1},t_L]$, where the discrete event times $0<t_1<t_2<...<t_L$. Define the $l$-th interval as $A_l=(t_{l-1},t_l]$ and $T_i=l$ means that the event occurred in the time interval $A_l$. Under this definition, the discrete probability function can be expressed as:
\begin{equation*}
    f_l=P(T\in A_l)=S(t_{l-1})-S(t),
\end{equation*}
where $S(t_l)=P(T>t_l)$ represents the survival function. Furthermore, the discrete hazard rate $h_l$ is defined as the conditional failure probability:
\begin{equation*}
    h_l=P(T\in A_l|T>t_{l-1})=\frac{f_l}{S(t_{l-1})}.
\end{equation*}
When introducing the discrete-time censoring indicator $\delta_{il}$, the total likelihood can be expressed as:
\begin{equation}
\ell=\prod_{i=1}^N \prod_{l=1}^{l_i}h_{il}^{\delta_{il}}(1-h_{il})^{1-\delta_{il}} =\prod_{l=1}^L \prod_{i\in R_l}h_{il}^{\delta_{il}}(1-h_{il})^{1-\delta_{il}},
\label{eq:ds-like}
\end{equation}
where $R_l$ is the set of individual $i$ at risk in the $l$-th interval of time period. In \autoref{eq:ds-like}, the indicator $\delta_{il}$ equals 1 in the interval $A_l$ containing the event of interest for the uncensored subjects and equals 0 otherwise.

\subsection{Discrete-time Cause-specific Model}
One of the popular approaches to modeling discrete-time competing-risks data is the discrete cause-specific hazards model. It involves the CIF to a set of covariates for the cause-specific hazards $\lambda_m(t|x_i)$:
\begin{equation*}
    \lambda_m(t|x_i)=P(T_i=t,R_i=m|T_i\geq t,x_i).
\end{equation*}
Based on the cause-specific hazard function, the CIF for the $m$-th event can be expressed as:
\begin{equation*}
    F_m(t|x_i)=\sum_{s=1}^t\lambda_m(s|x_i)S(s-1|x_i),
    \label{eq:cif-cs}
\end{equation*}
where $S(t|x_i):=P(T_i>t|x_i)$ is the survival function denoting the probability of experiencing the first event after $t$.

For traditional modeling, the cause-specific hazards are estimated by specifying a multinomial logistic regression model defined as:
\begin{equation}
    \lambda_m(t|x_i)=\frac{\mathrm{exp}(\gamma_{0tm} + x_i^{\top}\gamma_{m})}{1+\sum_{j=1}^M \mathrm{exp}(\gamma_{0tj} + x_i^{\top}\gamma_{j})}, \text{for}\ m=1,...,M, t=1,2,...,
    \label{eq:hazards-cs}
\end{equation}
where $\gamma_{0tm}$ represents the cause-specific baseline coefficient for enent type $m$ at time $t$ and $\gamma_{m}$ is the vector of regression coefficients for event type $m$. The denominator includes all possible events plus the category of surviving beyond time $t$.
In most applications, this refers to “staying at risk”. To use multinomial logit models in practice, the data have to be reformatted in a way that for a subject with observed event time $t_i$, and can be defined as:
\begin{equation*}
    y_{is}=(y_{is0},y_{is1},...,y_{isM})=(1,0,...,0), \text{for all}\ s=1,...,t_i-1.
\end{equation*}
If the event $R_i$ is observed for observation $i$ at time $t_i$ (where $\delta_i=1$), we define:
\begin{equation*}
    y_{it_{i}}=(y_{it_{i}0},y_{it_{i}1},...,y_{it_{i}M})=(0,...,1,...,0),
\end{equation*}
with $y_{it_{i}R_{i}}=1$ and all other elements of $y_{it_{i}}$ set to zero. If subject $i$ is censored at time $t_i$ (where $R_i=0$ and $\delta_i=0$), we define:
\begin{equation*}
    y_{it_{i}}=(y_{it_{i}0},y_{it_{i}1},...,y_{it_{i}M})=(1,0,...,0).
\end{equation*}
This data structure is termed as the augmented set of observations according to \citep{berger2018semiparametric} and can be summarized as:
\begin{equation}
    y_{i{t_i}}^\top=(y_{it_{i}0},y_{it_{i}1},...,y_{i{\Tilde{T}_i}{R_i}},...,y_{it_{i}M})=
    \begin{cases}
        &(1,0,...,0,...,0),\ \text{if}\ t<\Tilde{T}_i,\\
        &(0,0,...,1,...,0),\ \text{if}\ t=\Tilde{T}_i,\ \delta_i=1,\\
        &(1,0,...,0,...,0),\ \text{if}\ t=\Tilde{T}_i,\ \delta_i=0.\\
    \end{cases}\label{eq:cs-aug}
\end{equation}
When the term $y_{i{\Tilde{T}_i}{R_i}}=1$ in the second row of \autoref{eq:cs-aug}, the log-likelihood can be written as:
\begin{equation*}
    \ell=\sum_{i=1}^N \sum_{t=1}^{\Tilde{T}_i}\{\sum_{m=1}^M y_{itm}\mathrm{log}(\lambda_m(t|x_i))+y_{it0}\mathrm{log}(1-\lambda(t|x_i))\},
\end{equation*}
where $\lambda(t|x_i):=P(T_i=t|T_i\geq t,x_i)=\sum_{m=1}^M\lambda_m(t|x_i)$ is the overall hazard function representing the probability of experiencing any of the $M$ events at $t$.

\subsection{Discrete-time Sub-distribution Model}
A disadvantage of the discrete cause-specific hazards model defined in \autoref{eq:hazards-cs} is the lack of a one-to-one relationship between $\lambda_m(t|x_i)$ and $F_m(t|x_i)$ according to \citep{schmid2021competing}. In reality, \autoref{eq:cif-cs} suggests that it's necessary to model all $M$ cause-specific hazard functions in order to determine the CIF $F_m(t|x_i)$ for the $m$-th event. This becomes problematic when the focus is solely on one of the $M$ events and when there is a large number of competing events, leading to an extensive number of coefficients to estimate.

To tackle this limitation, an alternative approach called the discrete sub-distribution hazard model is developed by \citep{berger2020subdistribution}, which directly models the CIF of one specific event of interest. This method builds upon the modeling approach for continuous event times developed by \citep{fine1999proportional}. A key benefit is that it allows for interpreting covariate effects on event occurrence through a single model, simplifying the analysis.

For simplicity, we will discuss the discrete-time Sub-distribution model by focusing on modeling the occurrence of a type 1 event ($R_i=1$), taking into account that there are $M-1$ competing events and also the censoring event ($\delta_i=0$). The other events can be modeled with similar representations. We estimate the CIF for a type 1 event as:
\begin{equation*}
    F_1(t|x_i)=P(T_i\leq t, R_i=1|x_i).
\end{equation*}
Based on the assumption that the type 1 event will never be the first event to be observed once a competing event has occurred, implying that there is no finite event time for the occurrence of a type 1 event if $R_i\neq 1$ made by \citep{fine1999proportional}, the discrete sub-distribution time of the type 1 event is defined as:
\begin{equation*}
    \vartheta_i:=
    \begin{cases}
        T_i,\ \text{if}\ R_i=1,\\
        \infty,\ \text{if}\ R_i\neq 1.
    \end{cases}
\end{equation*}
Accordingly, the discrete sub-distribution hazard function is defined as:
\begin{equation*}
\begin{aligned}
    \xi_1(t|x_i)&=P(T_i=t,R_i=1|(T_i\geq t)\cup (T_i\leq t-1,R_i\neq 1),x_i)\\
    &=P(\vartheta_i=t|\vartheta_i \geq t,x_i).
\end{aligned}
\end{equation*}
The relationship between the discrete sub-distribution hazard $\xi_1$ and the CIF $F_1$ can be derived as:
\begin{equation}
    F_1(t|x_i)=1-\prod_{s=1}^t(1-\xi_1(s|x_i))=1-S_1(t|x_i),
    \label{eq:cif-f1}
\end{equation}
where $S_1(t|x_i)=P(\vartheta_i \geq t|x_i)$ is the discrete survival function for a type 1 event. \autoref{eq:cif-f1} implies that a statistical model for the discrete sub-distribution hazard has a direct interpretation in terms of the cumulative incidence function. A classical parametric model for $\xi_1(t|x_i)$ is given as:
\begin{equation}
    \xi_1(t|x_i)=\frac{\mathrm{exp}(\beta_{0t1} + x_i^\top\beta_1)}{1 + \mathrm{exp}(\beta_{0t1}+x_i^\top\beta_1)},
    \label{eq:hazards-sub}
\end{equation}
where $\beta_{0t1}$ are the parameters representing the sub-distribution baseline coefficients for event type 1 at time $t$ and $\beta_1$ are the regression parameters for event type 1. The transformed data for a subject with observed event time $t_i$ is defined as:
\begin{equation*}
    (y_{i1},...,y_{i,{\Tilde{T}_i}},...,y_{i,{L-1}})=
    \begin{cases}
        (0,...,0,1,0,...,0),\ \text{if}\ \delta_i R_i=1,\\
        (0,...,0,0,0,...,0),\ \text{if}\ \delta_i R_i\neq 1.
    \end{cases}
\end{equation*}
With this reformatted data, the total weighted log-likelihood shown in \citep{berger2020subdistribution} can be written as:
\begin{equation*}
    \ell=\sum_{i=1}^N\sum_{t=1}^{L-1} w_{it}\{y_{it}\mathrm{log}(\xi_1(t|x_i))+(1-y_{it})\mathrm{log}(1-\xi_1(t|x_i))\},
\end{equation*}
where $w_{it}=I(t\leq \Tilde{T}_i)$ is a set of weights indicating whether subject $i$ is at risk or not at time $t$. The weights are estimated based on the censoring survival function $\hat{G}(t)=\hat{P}(C_i>t)$ given as:
\begin{equation*}
    w_{it}=\frac{\hat{G}(t-1)}{\hat{G}(\mathrm{min}(\Tilde{T}_i,t)-1)}\cdot(I(t\leq \Tilde{T}_i)+I(\Tilde{T}_i\leq t-1,\delta_i R_i>1)).
\end{equation*}

\subsection{Deep-learning Framework with Functional Covariates and Missing Data}

\begin{figure}[t]
    \centering
    \includegraphics[width=0.9\linewidth]{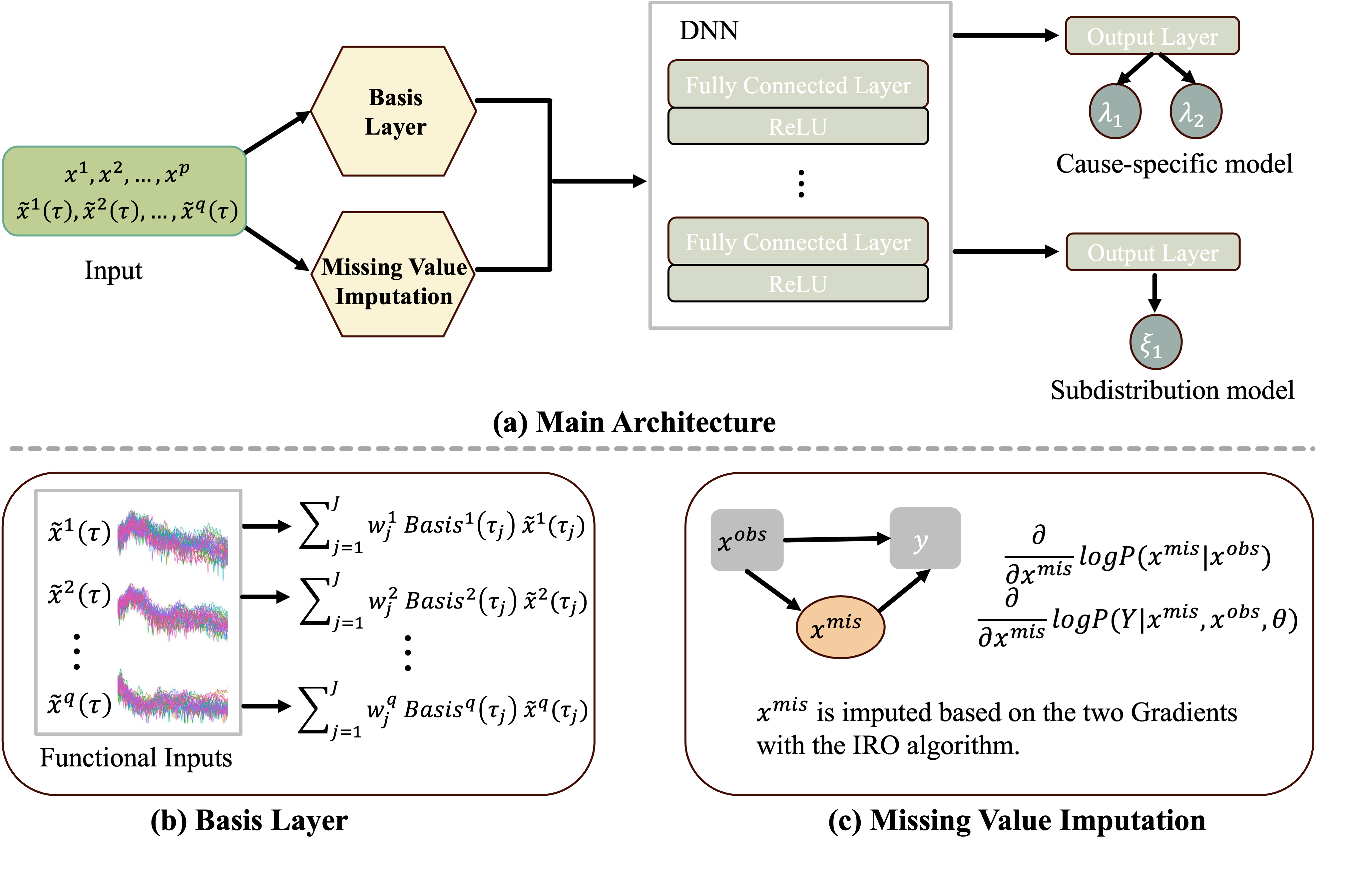}
    \caption{The overall deep-learning framework of our method. (a) The main network architecture consists of three components for the Cause-specific and the sub-distribution model. (b) Basis layer utilizes the micro neural network with functional input. (c) Algorithm of missing value imputation based on the observed data and the network outcome $y$.}
    \label{fig:structure}
\end{figure}

We propose a novel deep-learning-based method to estimate the two different hazard functions, incorporating multi-dimensional functional variables, multi-dimensional static variables, and simultaneously handling missing data in the ICU study. In \autoref{fig:structure}, we show the overall architecture of our model. It consists of three components: Basis Layer, Missing Value Imputation, Fully Connected Layer.

In the main network architecture, we develop two distinct types of DNNs to estimate the cause-specific hazard function and the sub-distribution hazard function. This is achieved by creating different output layers tailored to the specifications detailed in \autoref{eq:hazards-cs} and \autoref{eq:hazards-sub}. Given the paired data structure including the covariate vector $x_i$ and functional data $\tilde{x}_i(\tau)$, we aim to predict the discrete time-to-event outcome $\hat{y}_i=\mathcal{F}((x_i,\tilde{x}_i(\tau));\theta)$, where $\mathcal{F}$ is the designed network and $\theta$ are the network parameters to be learned. These DNNs are specifically designed to handle the complexities of competing risks in survival analysis, where different types of events can preclude the occurrence of the event of interest. Additionally, our DNN framework addresses the challenges associated with handling functional data and managing missing data, which are prevalent issues in many real-world datasets. 
Our network design incorporates specialized layers that can process and integrate these complex data types into the survival analysis, enhancing the model’s ability to leverage the rich information contained within functional data. 

For the main DNNs, we leverage the fully connected (FC) layer combined with $\mathrm{ReLU()}$ activation layer to process the input data expressed as:
\begin{equation}
    \begin{aligned}
        &z=\mathrm{MVI}(x,\mathrm{BL}(\tilde{x})),\\
        &h_1=\mathrm{FC}_1(z),\\
        &h_i=\mathrm{ReLU}(\mathrm{FC}_i(h_{i-1})),\ i=2,3,...,n_h\\
        &\hat{y}=\mathrm{Output}(h_{n_h}),
    \end{aligned}
\end{equation}
where $h$ is the hidden state and $n_h$ denotes the number of hidden layers. $z$ is the transformed input features by the designed two pre-processing modules. According to the preliminary in Sections 3.2 and 3.3, the log-likelihood can be equivalent to the cross-entropy loss for classification when estimated with neural networks. The total loss function of our model for the cause-specific model and sub-distribution model can be written as:
\begin{equation}
    \mathcal{L}_{cs} = -\sum_{i=1}^N y_i\mathrm{log}(\hat{y}_i).\label{eq:loss-cs}
\end{equation}

\begin{equation}
    \mathcal{L}_{sub} = -\sum_{i=1}^N w_i y_i\mathrm{log}(\hat{y}_i).\label{eq:loss-sub}
\end{equation}
$y_i$ are the reformatted long-form data. Our aim is to minimize the loss function to better predict the time-to-event outcome.

\begin{figure}[t]
    \centering
    \includegraphics[width=0.8\linewidth]{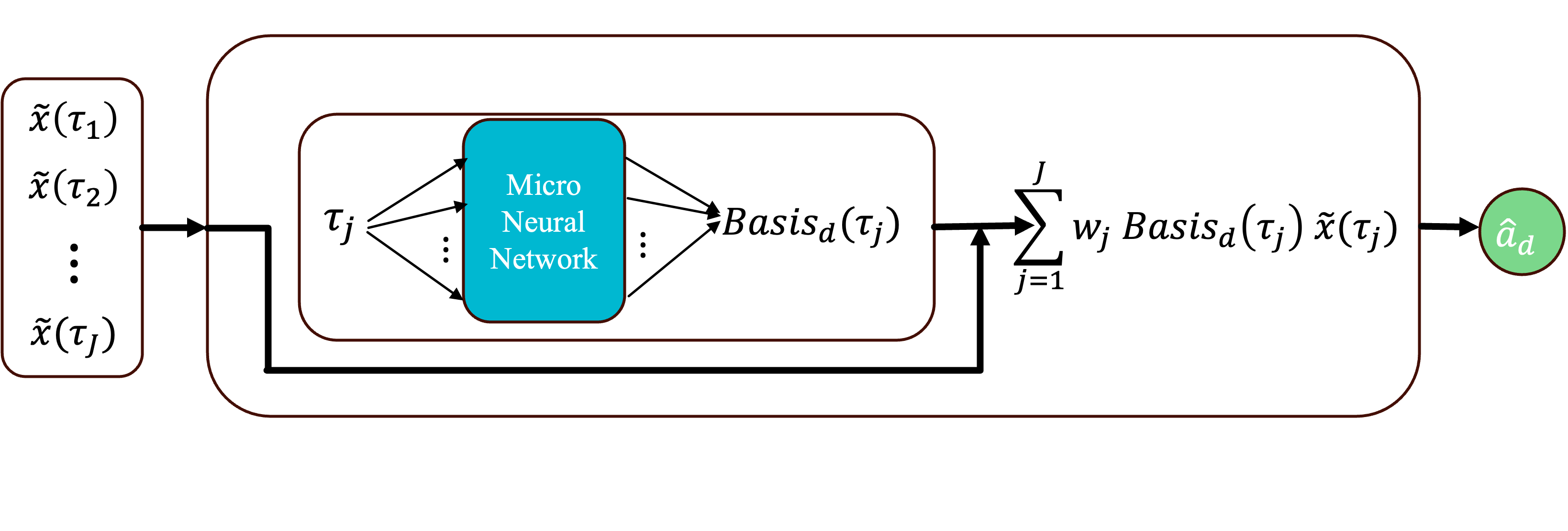}
    \caption{Detailed structure of the Basis Layer.}
    \label{fig:basis-layer}
\end{figure}

To flexibly represent functional variables within a deep learning framework, we adapt \citet{yao2021deep}'s micro-network idea. They proposed to replace the fixed basis functions $\phi_d(\tau)$ in Equation (\ref{eq:fda}) with an adaptive Basis Layer consisting of $D$ basis nodes, in which each basis node represents the adaptive basis function $B_d(\tau)$, for $d=1,...,D$ and $\tau\in[0,1]$, and is parameterized by a small neural network—referred to as a micro-network. The form of a functional variable remains:
$$
\tilde{x}(\tau) \approx \sum_{d=1}^D a_d \, B_d(\tau).
$$
In our model, a vector $\bm{\mathrm{v}}_a=[a_1,...,a_D]\in \mathbb{R}^D$, where
\begin{equation*}
    a_d = \int \tilde{x}(\tau) B_d(\tau) d\tau,
\end{equation*}
is generated from the micro-network and serves as the input for the main networks to represent the functional variable. The coefficient $a_d$ serves as the functional projection of $\tilde{x}(\tau)$ onto the learned basis $B_d(\tau)$, analogous to classical basis expansion coefficients. Unlike traditional FDA, where basis functions are pre-specified (e.g., splines, Fourier), the adaptive Basis Layer allows basis functions to be learned directly from data, enabling the network to capture complex, nonlinear patterns in functional signals in a compact and data-adaptive manner.
These basis layers are implemented as a micro-network trained simultaneously with the main network, as shown in \autoref{fig:basis-layer}. Each basis function $B_d(\tau)$ is parameterized by a micro-network $Basis_d(\tau)$ for the $d$-th basis node, where $\tau$ is the time point of the functional data. 
The micro network maps the input $\tau$ to $B_d(\tau)$ via neural network transformations:
\begin{equation*}
    B_d(\tau)=Basis_d(\tau)=\sigma_K^d(\dots \sigma_1^d(W_1^d\tau +b_1^d)),
\end{equation*}
where $\{W_k^d,b_k^d\}_{k=1}^K$ are the weights and biases of the micro network and $\sigma_k$ are activation functions at each sub-layer. The $d$-th vector representation $a_d$ is the inner product between the basis function and the functional input $\tilde{x}(\tau)$, which is estimated as:
\begin{equation*}
    \hat{a}_d=\sum_{j=1}^{J}w_j B_d(\tau_j)\tilde{x}(\tau_j)=\sum_{j=1}^{J}w_j Basis_d(\tau_j)\tilde{x}(\tau_j),
\end{equation*}
where $\tau_j$ are discrete sample points of $\tau$ of the functional covariates and $w_j$ are weights from numerical integration methods.

In many real-world datasets, a significant number of missing values are often present. This issue can complicate data analysis and affect the accuracy of conclusions drawn from the data. Traditional imputation methods may not perform well in high-dimensional settings due to computational complexity and inability to capture complex dependencies among variables \citep{sun2023deep}. Inspired by the Imputation-regularized optimization (IRO) algorithm proposed by \citep{liang2018imputation}, which aims to address missing data problems in high-dimensional contexts, we integrate the IRO algorithm into our deep learning framework to simultaneously perform missing data imputation and prediction. Given the covariate vector $x=(x_1,x_2,...,x_N)$ with potential missing values, each $x_i$ of subject $i$ can be divided into observed parts $x_i^{obs}$ and missing parts $x_i^{mis}$. We can assume $x^{mis}\perp \mathbf{R}|(x^{obs},y)$, meaning that the missing values are conditionally independent of $\mathbf{R}$ given the observed data and the outcome $y$. The joint distribution can be expressed as:
\begin{equation*}
    P(x^{mis},y|x^{obs},\mathbf{R},\theta)\propto P(x^{miss}|x^{obs})P(y|x^{obs},x^{mis},\theta),
\end{equation*}
where $\theta$ are the network parameters to be optimized and $\mathbf{R}$ is the missingness indicator. The missing parts $x^{mis}$ will be imputed based on the conditional probability of $P(x^{mis}|x^{obs})$ and $P(y|x^{obs},x^{mis},\theta)$. In the training stage, this operation will be automatically achieved during network learning according to the two gradients on $x^{mis}$:
\begin{equation*}
\begin{aligned}
    \frac{\partial}{\partial x^{mis}}\mathrm{log}P(x^{mis}|x^{obs}),\\
    \frac{\partial}{\partial x^{mis}}\mathrm{log}P(y|x^{obs},x^{mis},\theta).
\end{aligned}
\end{equation*}

The first gradient encourages imputations that are plausible given the observed data, and the second one encourages imputations that improve the predictive performance of the model. In the modeling of $P(x^{mis}|x^{obs})$, we leverage the Gaussian graphical network (GGN) to capture dependencies among variables in the data. GGN represents variables as nodes in a graph, with edges indicating conditional dependencies. For a missing entry $x_{ij}$ denoting the $j$-th covariate of subject $i$, let $\omega(j)={p:a_{jp}=1}$ denote the neighborhood of covariate $j$, where $a_{jp}$ are entries in the adjacency matrix defining the graph structure. Conditionally on the neighborhood $\omega(j)$, $x_{ij}$ can be imputed on the expression values of the neighboring covariates:
\begin{equation*}
    \left(\begin{matrix}
        x_{ij}\\ x_{i\omega(j)}
    \end{matrix} \right)
    \sim \mathcal{N}
    \left( \left(\begin{matrix}
        \mu_j\\ \mu_{\omega(j)}
    \end{matrix}\right),
    \left(\begin{matrix}
        \sigma_j^2 & \Sigma_{j\omega(j)}\\ \Sigma_{j\omega(j)}^\top & \Sigma_{\omega(j)\omega(j)}
    \end{matrix}\right) \right),
\end{equation*}
where $x_{i\omega(j)}$ are the observed values of neighboring variables and $\mu_j,\mu_{\omega(j)},\sigma_j^2,$ $\Sigma_{j\omega(j)},\Sigma_{\omega(j)\omega(j)}$ are the corresponding mean and variance components to be estimated from the data. In this way, we have the following IRO algorithm for learning a Gaussian graphical model in the presence of missing data. We also introduce a stochastic modification to the implied IRO algorithm to enhance its ability to explore the parameter space more effectively.
\begin{itemize}
    \item \textbf{Initialization}: Fill each missing entry with initial estimates, such as the median of the observed values.
    \item \textbf{I-Step (Imputation)}: Update $x^{mis}$ using a stochastic gradient Langevin dynamics (SGLD) incorporating both gradients as follows:
    \begin{equation*}
    \small
          x^{mis}\gets x^{mis}+\eta \left( \frac{\partial}{\partial x^{mis}}\mathrm{log}P(x^{mis}|x^{obs}) + \frac{\partial}{\partial x^{mis}}\mathrm{log}P(y|x^{obs},x^{mis},\theta) \right)+ \sqrt{2 \eta} {\bf e}, 
    \end{equation*}
    where ${\bf e}\sim N(0,I_{d_{x^{mis}}})$, ${d_{x^{mis}}}$ is the dimension of $x^{mis}$, and $\eta$ is the updating learning rate for imputation.
    \item \textbf{RO-Step (Parameter Update)}: Update network parameters $\theta$ by minimizing the loss function \autoref{eq:loss-cs} or \autoref{eq:loss-sub} and learn the structure of the Gaussian graphical network between $x^{mis}$ and $x^{obs}$ to find the minimal divergence.
    \item \textbf{Convergence}: Iterate the I-Step and RO-Step until convergence criteria are met.
\end{itemize}


\section{Experimental Results}
In this section, we evaluate the proposed FCRN through a series of experiments on both synthetic and real-world datasets. 
We design two scenarios with and without missing data to validate the effectiveness of our framework.
We compare FCRN against established baselines, including Random Survival Forest (RSF) and traditional Competing-risks Regression (CR) models. Performance is quantified using the Integrated Brier Score (IBS), reported separately for two competing event causes (Cause 1 and Cause 2). For synthetic data, we report the means and standard deviations from 100 independent repetitions to ensure statistical reliability.

\subsection{Dataset}

\noindent\textbf{Synthetic Data}
We generated synthetic datasets under two parallel situations: one with only tabular covariates, and another that also included functional covariates. Each situation has $n=100$ datasets, and each dataset contains $n=1,000$ subjects. The tabular data includes 10 base covariates: five from normal distributions (e.g., $N(0.2,1)$, $N(1.5,1.2)$), and five from uniform distributions (e.g., $\mathrm{Uniform}[-1,1]$, $\mathrm{Uniform}[0.2,2.0]$). To introduce complex, latent structures, an additional five covariates capturing nonlinearities and interactions (e.g., $\text{norm}_1^2$) were used in the outcome generation process but were subsequently removed from the final datasets provided to the models. The datasets containing functional covariates were constructed by augmenting the 10 base tabular covariates with three functional predictors per subject. Each functional predictor was generated using a B-spline basis with 15 basis functions, evaluated over 51 discrete time points.

Competing risks outcomes (survival time and event indicator) for both datasets were generated using \texttt{crisk.ncens.sim} from the \texttt{cmprsk} R package, with lognormal and Weibull baseline distributions for the two event types. Individual-specific random effects, drawn from Gamma and Exponential distributions, were also incorporated. To assess robustness to missingness, we imposed missing-at-random (MAR) patterns, removing 25\% and 50\% of the values from 10 base tabular covariates. Finally, each of the 100 datasets was randomly split into training (800 subjects) and test (200 subjects) sets.

\begin{table}[t]
\caption{Statistical summary of the two real-world datasets.}
\centering
\label{tab:summary}
\begin{tabular}{cc|ccc}
\toprule\hline
                                    &           & \begin{tabular}[c]{@{}c@{}}MIMIC\\ Train set\end{tabular} & \begin{tabular}[c]{@{}c@{}}MIMIC\\ Test set\end{tabular} & \begin{tabular}[c]{@{}c@{}}CC\\ data\end{tabular} \\ \hline
\multirow{2}{*}{Age}                & Mean      & 64              & 65             & 65       \\
                                    & Std       & 15              & 14             & 16       \\
\multirow{2}{*}{BMI}                & Mean      & 29.16           & 29.28          & 29.26    \\
                                    & Std       & 7.40            & 7.89           & 8.30     \\
\multirow{2}{*}{ICU length of stay} & Mean      & 5.89            & 5.66           & 3.41     \\
                                    & Std       & 7.59            & 6.89           & 3.39     \\
\multirow{2}{*}{CCI}                & Mean      & 4.5             & 4.5            & 6.7      \\
                                    & Std       & 2.6             & 2.5            & 3.6      \\ \hline \hline
\multirow{2}{*}{Sex}                & Famle     & 36\%            & 36\%           & 55\%     \\
                                    & Male      & 64\%            & 64\%           & 45\%     \\
\multirow{3}{*}{Race}               & White     & 70\%            & 70\%           & 73\%     \\
                                    & Black     & 6\%             & 5\%            & 18\%     \\
                                    & Other     & 24\%            & 25\%           & 9\%      \\
\multirow{3}{*}{Admission Type}     & Emergency & 60\%            & 58\%           & 76\%     \\
                                    & Urgent    & 23\%            & 24\%           & 6\%      \\
                                    & Other     & 17\%            & 18\%           & 18\%     \\ \hline\bottomrule
\end{tabular}
\end{table}

\noindent\textbf{MIMIC Data}
MIMIC-IV is a large, freely available database comprising de-identified health-related data associated with over 65,000 patients admitted to an ICU and over 200,000 patients admitted to the emergency department at the Beth Israel Deaconess Medical Center in Boston, MA \citep{johnson2020mimic,johnson2023mimic}. MIMIC-IV incorporates contemporary data and adopts a modular approach to data organization, highlighting data provenance and facilitating both individual and combined use of disparate data sources. This database is sourced from two in-hospital database systems: a custom hospital-wide EHR and an ICU-specific clinical information system.
The database includes information such as demographics, vital sign measurements made at the bedside, laboratory test results, procedures, medications, caregiver notes, imaging reports, and mortality. All patient data has been meticulously de-identified to maintain confidentiality and comply with privacy regulations. 

\begin{figure}
    \centering
    \includegraphics[width=0.9\linewidth]{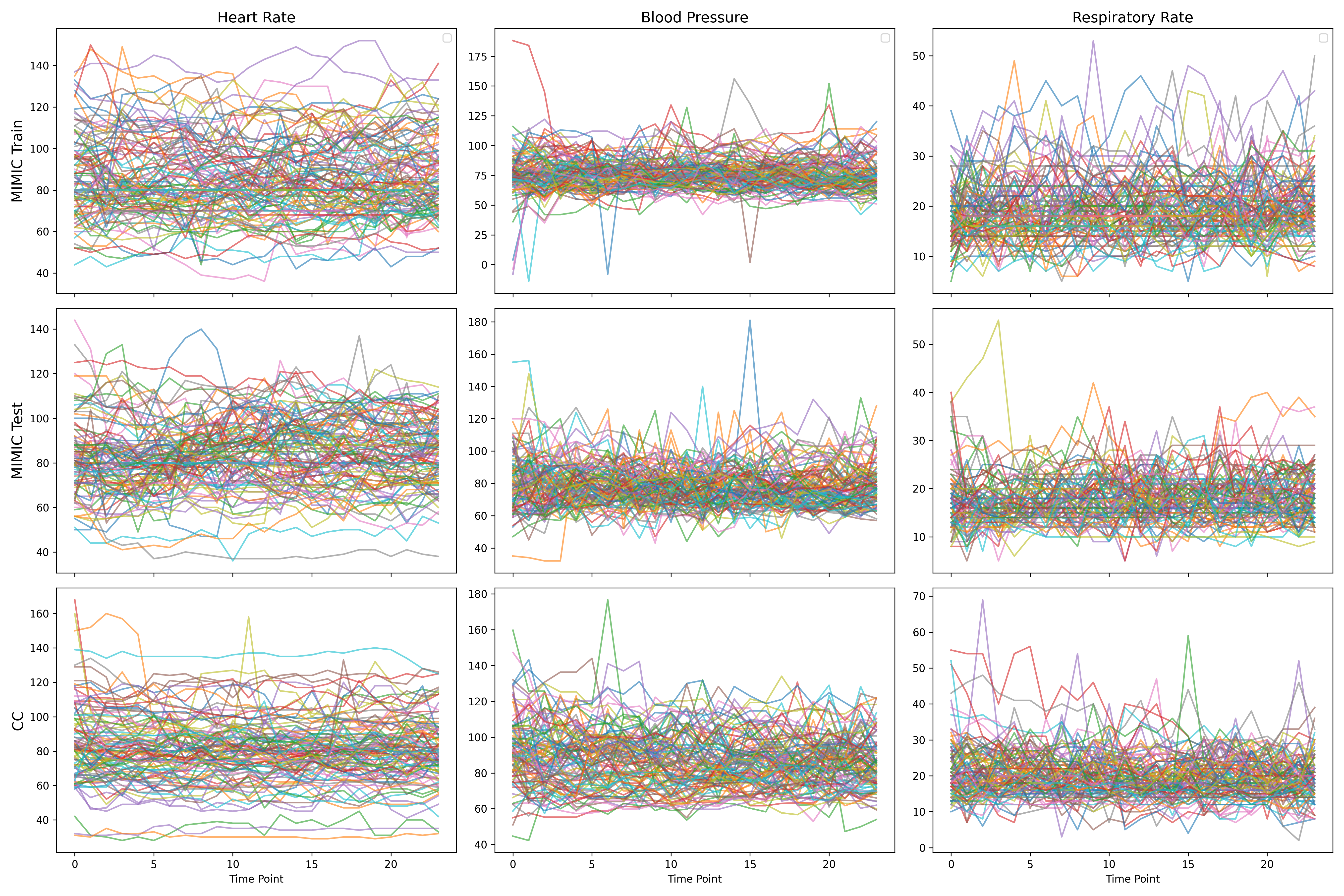}
    \caption{Sampled functional curves for the two real-world datasets.}
    \label{fig:func}
\end{figure}

\noindent\textbf{CC Data}
We retrospectively collected data from patients admitted to the intensive care units (ICUs) of the Cleveland Clinic (CC) between January and December 2023. The study was approved by the CC Institutional Review Board (IRB 14-1431 and 20-404), and the requirement for written informed consent was waived. All patient data were de-identified and systematically captured through standardized clinical templates within a proprietary data repository implemented across the health care system. The dataset included clinical information extracted from the electronic medical records of all patients aged 18 years or older who were admitted to any ICU within the CC Health System in Ohio, irrespective of the cause of admission. Patients were excluded if they died or were discharged on the day of ICU admission. Demographic variables, including age, sex, race, body mass index (BMI), and Charlson Comorbidity Index (CCI), were collected. Admission-related information, such as the type of admission (inpatient, emergency department, urgent care, general floor, or other), was also retrieved. 
Vital signs, including blood pressure, heart rate, and respiratory rate, were recorded approximately hourly throughout the ICU stay as functional data input. \autoref{tab:summary} shows the statistical summary of the two real-world datasets for numerical variables and categorical variables. \autoref{fig:func} presents the functional curves of three vital signs for a randomly selected subset of 100 patients. 

\subsection{Evaluation Metrics}
We use the IBS to assess the predictive accuracy of survival functions over specified horizons. Baseline methods model competing risks in continuous time, while FCRN employs a discrete-time approach; thus, we avoid the concordance index due to potential inconsistencies across frameworks.

  
\noindent\textbf{Integrated Brier Score}
The Brier score (BS) quantifies the mean squared difference between observed survival outcomes and predicted probabilities, which is defined as 

\[
\mathrm{BS}(t) = \mathbb{E}_{x \sim} \left[ \| \mathbf{1}\{T > t\} - \hat{\mathbb{P}}(T > t \mid X) \|_2^2 \right]
\]
A lower Brier score indicates a better model prediction at the determined time point. Over a specified time interval, the integrated Brier score is obtained by averaging the Brier score over that interval. In practice, this is done by calculating the area under the curve of the score plotted against follow-up time and then dividing by the length of the interval. The Integrated Brier Score for censored observation between the time interval $[t_0, t_{\max}]$ is defined as
\[
  \mathrm{IBS}^c = \frac{1}{t_{\max} - t_0}
  \int_{t_0}^{t_{\max}} \mathrm{BS}^c(t)\,\mathrm{d}t,
\]
where a lower Integrated Brier score also indicates a better model prediction.

\subsection{Implementation Details}
We implement our framework using the open-source PyTorch equipped with an NVIDIA V100 GPU. The number of layers of our main network is set as 3, and the hidden units are set as [32, 64, 32] correspondingly. Adam \citep{kingma2014adam} has been used as an optimizer to minimize the objective function with an initial learning rate of
0.001. For the missing value imputation, we set the impute learning rate at 0.003 with a step size of 0.1. The batch size is set as 64. In the Basis Layer, the number of basis functions is grid searched from 2 to 8. The length of the discrete time interval is set to 5 across the synthetic, MIMIC, and CC datasets, resulting in 20 intervals for the synthetic dataset, and [37, 73] intervals of half/one year prediction for the real-world datasets.

\begin{table}[t]
    \centering
    \caption{{Performance comparison of Random Survival Forest (RSF), traditional Competing-risk Regression model (CR), and our proposed FCRN across complete and missing data scenarios with IBS metrics as evaluation.} CSM: Cause-specific model; SDM: Sub-distribution model.}
    \label{tab:miss_sys}
    \setlength{\tabcolsep}{1mm}
    \resizebox{0.99\textwidth}{!}{
    \begin{tabular}{c p{3.2cm} ccccc} 
    \toprule \hline
        ~ & ~ & ~ & \multicolumn{4}{c}{IBS} \\ \cline{4-7}
       ~ & ~ & ~ & \multicolumn{2}{c}{Cause 1} & \multicolumn{2}{c}{Cause 2} \\ \cline{4-7}
        ~ & ~ & ~ & CSM & SDM & CSM & SDM \\ \hline \hline
        \multicolumn{1}{c|}{\multirow{6}{4.5cm}{\centering Original Training Data and Testing Data (800 subjects vs 200 subjects)}} 
        & \multirow{2}{3.5cm}{\centering RSF}
        & Mean & 0.167 & 0.160 & 0.194 & 0.185 \\ 
        \multicolumn{1}{c|}{~} & ~ & Std & 0.011 & 0.011 & 0.008 & 0.008 \\ \cline{2-7}
        \multicolumn{1}{c|}{~} & \multirow{2}{3.5cm}{\centering CR}
        & Mean & 0.185 & 0.175 & 0.220 & 0.213 \\ 
        \multicolumn{1}{c|}{~} & ~ & Std & 0.016 & 0.013 & 0.011 & 0.010 \\ \cline{2-7}
        \multicolumn{1}{c|}{~} & \multirow{2}{3.5cm}{\centering FCRN} 
        & Mean & 0.153 & 0.157 & 0.176 & 0.188 \\ 
        \multicolumn{1}{c|}{~} & ~ & Std & 0.013 & 0.013 & 0.014 & 0.015 \\ \hline
        \multicolumn{1}{c|}{\multirow{6}{4.5cm}{\centering 25\% Missingness of Entire Data (MAR)}} 
        & \multirow{2}{3.5cm}{\centering RSF with median imputation}
        & Mean & 0.178 & 0.175 & 0.207 & 0.203 \\ 
        \multicolumn{1}{c|}{~} & ~ & Std & 0.012 & 0.012 & 0.009 & 0.010 \\ \cline{2-7}
        \multicolumn{1}{c|}{~} & \multirow{2}{3.5cm}{\centering CR with median imputation} 
        & Mean & 0.189 & 0.183 & 0.225 & 0.221 \\ 
        \multicolumn{1}{c|}{~} & ~ & Std & 0.015 & 0.014 & 0.011 & 0.010 \\ \cline{2-7}
        \multicolumn{1}{c|}{~} & \multirow{2}{3.5cm}{\centering FCRN with MVI} 
        & Mean & 0.173 & 0.172 & 0.201 & 0.203 \\ 
        \multicolumn{1}{c|}{~} & ~ & Std & 0.014 & 0.014 & 0.014 & 0.014 \\ \hline
        \multicolumn{1}{c|}{\multirow{6}{4.5cm}{\centering 50\% Missingness of Entire Data (MAR)}} 
        & \multirow{2}{3.5cm}{\centering RSF with median imputation} 
        & Mean & 0.197 & 0.196 & 0.223 & 0.222 \\ 
        \multicolumn{1}{c|}{~} & ~ & Std & 0.013 & 0.014 & 0.009 & 0.010 \\ \cline{2-7}
        \multicolumn{1}{c|}{~} & \multirow{2}{3.5cm}{\centering CR with median imputation} 
        & Mean & 0.198 & 0.194 & 0.232 & 0.231 \\ 
        \multicolumn{1}{c|}{~} & ~ & Std & 0.015 & 0.015 & 0.010 & 0.009 \\ \cline{2-7}
        \multicolumn{1}{c|}{~} & \multirow{2}{3.5cm}{\centering FCRN with MVI} 
        & Mean & 0.189 & 0.188 & 0.218 & 0.217 \\ 
        \multicolumn{1}{c|}{~} & ~ & Std & 0.014 & 0.014 & 0.010 & 0.010 \\ \hline
    \bottomrule
    \end{tabular}
    }
\end{table}

\subsection{Results of Synthetic Data}

The quantitative results of only containing tabular covariates for complete and missing settings are shown in \autoref{tab:miss_sys}. The first panel shows the compared results for the complete settings. FCRN consistently demonstrates lower mean IBS values for both causes compared to RSF and CR, indicating superior predictive accuracy and better calibration. For instance, under Cause 1 with the CSM, FCRN attained a mean IBS of 0.153, compared to 0.167 for RSF and 0.185 for CR. Similar improvements were observed for SDM. 
The second and third panels show the compared results regarding the missing settings with 25\% and 50\% missing ratios for the entire dataset. Our FCRN method employing the MVI component continues to outperform other methods under 25\% missing ratio, achieving lower IBS values, such as 0.173 on CSM and 0.172 on SDM for Cause 1, and similarly favourable results for Cause 2. These results highlight FCRN’s robustness in handling incomplete data without substantial degradation in predictive power.
The superior performance of FCRN is further reinforced under higher missingness conditions, consistently yielding lower IBS values. For example, in the CSM setting of Cause 1, FCRN achieved a mean IBS of 0.189, lower than RSF (0.197) and CR (0.198). These findings demonstrate that FCRN not only surpasses established methods in complete-data scenarios but also exhibits resilience to high levels of missingness, preserving predictive reliability in challenging settings. 

\begin{table}[t]
    \centering
    \caption{{Performance comparison of Random Survival Forest (RSF), traditional Competing-risk Regression model (CR), and our proposed FCRN across complete and missing data scenarios when involving functional data with IBS metrics as evaluation.} CSM: Cause-specific Model; SDM: Sub-distribution model.}
    \label{tab:miss_sys_fda}
    \setlength{\tabcolsep}{1mm}
    \resizebox{0.99\textwidth}{!}{
    \begin{tabular}{c p{3.2cm} ccccc} 
    \toprule \hline
        ~ & ~ & ~ & \multicolumn{4}{c}{IBS} \\ \cline{4-7}
       ~ & ~ & ~ & \multicolumn{2}{c}{Cause 1} & \multicolumn{2}{c}{Cause 2} \\ \cline{4-7}
        ~ & ~ & ~ & CSM & SDM & CSM & SDM \\ \hline \hline
        \multicolumn{1}{c|}{\multirow{6}{4.5cm}{\centering Original Training Data and Testing Data (800 subjects vs 200 subjects)}} 
        & \multirow{2}{3.5cm}{\centering RSF}
        & Mean & 0.179 & 0.173 & 0.210 & 0.204 \\ 
        \multicolumn{1}{c|}{~} & ~ & Std & 0.010 & 0.010 & 0.006 & 0.006 \\ \cline{2-7}
        \multicolumn{1}{c|}{~} & \multirow{2}{3.5cm}{\centering CR}
        & Mean & 0.201 & 0.180 & 0.231 & 0.218 \\ 
        \multicolumn{1}{c|}{~} & ~ & Std & 0.022 & 0.014 & 0.016 & 0.011 \\ \cline{2-7}
        \multicolumn{1}{c|}{~} & \multirow{2}{3.5cm}{\centering FCRN} 
        & Mean & 0.161 & 0.158 & 0.187 & 0.192 \\ 
        \multicolumn{1}{c|}{~} & ~ & Std & 0.013 & 0.012 & 0.014 & 0.014 \\ \hline
        \multicolumn{1}{c|}{\multirow{6}{4.5cm}{\centering 25\% Missingness of Entire Data (MAR)}} 
        & \multirow{2}{3.5cm}{\centering RSF with median imputation}
        & Mean & 0.188 & 0.185 & 0.221 & 0.216 \\ 
        \multicolumn{1}{c|}{~} & ~ & Std & 0.011 & 0.011 & 0.006 & 0.007 \\ \cline{2-7}
        \multicolumn{1}{c|}{~} & \multirow{2}{3.5cm}{\centering CR with median imputation} 
        & Mean & 0.202 & 0.189 & 0.234 & 0.226 \\ 
        \multicolumn{1}{c|}{~} & ~ & Std & 0.019 & 0.014 & 0.015 & 0.012 \\ \cline{2-7}
        \multicolumn{1}{c|}{~} & \multirow{2}{3.5cm}{\centering FCRN with MVI} 
        & Mean & 0.179 & 0.174 & 0.210 & 0.211 \\ 
        \multicolumn{1}{c|}{~} & ~ & Std & 0.013 & 0.012 & 0.012 & 0.013 \\ \hline
        \multicolumn{1}{c|}{\multirow{6}{4.5cm}{\centering 50\% Missingness of Entire Data (MAR)}} 
        & \multirow{2}{3.5cm}{\centering RSF with median imputation} 
        & Mean & 0.199 & 0.198 & 0.231 & 0.228 \\ 
        \multicolumn{1}{c|}{~} & ~ & Std & 0.010 & 0.010 & 0.006 & 0.006 \\ \cline{2-7}
        \multicolumn{1}{c|}{~} & \multirow{2}{3.5cm}{\centering CR with median imputation} 
        & Mean & 0.207 & 0.199 & 0.240 & 0.236 \\ 
        \multicolumn{1}{c|}{~} & ~ & Std & 0.017 & 0.015 & 0.012 & 0.010 \\ \cline{2-7}
        \multicolumn{1}{c|}{~} & \multirow{2}{3.5cm}{\centering FCRN with MVI} 
        & Mean & 0.193 & 0.191 & 0.224 & 0.224 \\ 
        \multicolumn{1}{c|}{~} & ~ & Std & 0.014 & 0.014 & 0.009 & 0.010 \\ \hline
    \bottomrule
    \end{tabular}
    }
\end{table}

\autoref{tab:miss_sys_fda} shows the performance results when functional data are incorporated into the competing risk models, under both complete and missing data scenarios. Across the complete setting, FCRN consistently outperformed RSF and CR by yielding lower IBS values. For example, under Cause 1 with the CSM setting, FCRN achieved a mean IBS of 0.161 compared to 0.179 for RSF and 0.201 for CR. This advantage extended to Cause 2 as well. These results demonstrate the effectiveness of FCRN in leveraging functional data to improve risk prediction. Under both 25\% and 50\% missing ratios, the performance trends with functional data mirrored those observed without functional data. FCRN with MVI module consistently outperformed or matched RSF and CR with median imputation, demonstrating greater robustness to incomplete data. Although all methods showed degradation as missingness increased, FCRN maintained relatively stable predictive accuracy, underscoring its resilience across different data types.
Overall, the results highlight that incorporating functional data enhances the discriminative ability of FCRN compared to classical approaches, and exhibits resilience to varying levels of missingness, preserving competitive performance even under high proportions of incomplete data.

\subsection{Results of Real-world Data Analysis}

MIMIC data incorporate tabular variables (age, sex, BMI, CCI) and functional waveforms (heart rate, mean arterial blood pressure, respiratory rate). A test set of 2,000 subjects was randomly split; training used about 15,000 samples with missing values (missing mode) or about 11,000 after dropping incompletes (complete mode). CC data (about 20,000 patients) served as external validation. Competing events are ICU readmission (Cause 1) and death (Cause 2).

\begin{table}[t]
\centering
\caption{Comparison results of MIMIC data for Random Survival Forest (RSF), traditional Competing-risk Regression (CR), and our proposed FCRN. CSM: Cause-specific model; SDM: Sub-distribution model. Cause 1: ICU readmission; Cause 2: Death.}
\label{tab:mimic2_new}
\resizebox{0.8\textwidth}{!}{%
\begin{tabular}{ccccccccc}
\toprule\hline
 & \multicolumn{8}{c}{IBS (MIMIC dataset)} \\ \cline{2-9} 
 & \multicolumn{4}{c|}{Half year prediction} & \multicolumn{4}{c}{One year prediction} \\ \cline{2-9}
 & \multicolumn{2}{c}{Cause 1} & \multicolumn{2}{c|}{Cause 2} & \multicolumn{2}{c}{Cause 1} & \multicolumn{2}{c}{Cause 2} \\ \cline{2-9} 
 & CSM & SDM & CSM & \multicolumn{1}{c|}{SDM} & CSM & SDM & CSM & SDM \\ \hline\hline
\multicolumn{9}{c}{Scenario 1: Complete dataset with three functional variables} \\ \hline
\multicolumn{1}{c|}{RSF} & 0.190 & 0.190 & 0.030 & \multicolumn{1}{c|}{0.030} & 0.211 & 0.211 & 0.037 & 0.037 \\
\multicolumn{1}{c|}{CR} & 0.197 & 0.202  & 0.089 & \multicolumn{1}{c|}{0.101} & 0.215 & 0.224 & 0.107 & 0.122 \\
\multicolumn{1}{c|}{FCRN} & 0.175 & 0.176 & 0.028 & \multicolumn{1}{c|}{0.028} & 0.191 & 0.193 & 0.035 & 0.035 \\ \hline
\multicolumn{9}{c}{Scenario 2: Missing dataset with three functional variables} \\ \hline
\multicolumn{1}{c|}{RSF} & 0.185 & 0.186 & 0.033 & \multicolumn{1}{c|}{0.033} & 0.206 & 0.206 & 0.042 & 0.042 \\
\multicolumn{1}{c|}{CR} & 0.186 & 0.191 & 0.095 & \multicolumn{1}{c|}{0.103} & 0.204 & 0.208 & 0.101 & 0.108 \\
\multicolumn{1}{c|}{FCRN} & 0.176 & 0.176 & 0.031 & \multicolumn{1}{c|}{0.031} & 0.193 & 0.193 & 0.039 & 0.039 \\ \hline\bottomrule
\end{tabular}%
}
\end{table}

\autoref{tab:mimic2_new} presents a comparative result of competing risk prediction methods on the MIMIC dataset across two scenarios: (1) the complete dataset with three functional variables and (2) the dataset containing missing values with three functional variables. The evaluation metrics include Integrated Brier Score (IBS), assessed under both Cause-specific and Fine–Gray models at two time horizons of max follow-up time: half-year and one-year.
In the complete dataset, FCRN consistently achieves the lowest IBS across all causes and evaluation schemes for both half-year and one-year prediction horizons. For instance, in the half-year CSM prediction for Cause 1, FCRN records an IBS of 0.175, outperforming RSF (0.190) and CR (0.197). The trend persists in SDM and Cause 2 predictions, reflecting the robust calibration of FCRN.
The FCRN model maintains strong performance even under missing data. It achieves the lowest IBS values across most cases, such as 0.176 for half-year CSM of Cause 1 and 0.193 for one-year CSM of Cause 1, showing that FCRN is robust to incomplete information.
The FCRN method demonstrates superior performance on the MIMIC dataset through consistently lower IBS, and robustness under missing data. These results establish FCRN as a reliable and effective model for real-world survival analysis in the presence of competing risks.

In addition, we use the CC data as an external validation to evaluate the performance and generalization of FCRN. The results of CC data are tested on the model trained on the MIMIC data for corresponding scenarios without and with missing values. It is a crucial evaluation for assessing model robustness and clinical applicability. As shown in \autoref{tab:ccf_val}, FCRN achieves the lowest IBS across most cases for both half-year and one-year predictions. Particularly in CSM of Cause 1 and SDM of Cause 1 for ICU readmission, FCRN outperforms others by a large margin. 
Despite being trained on MIMIC data, FCRN demonstrates superior generalization to CC data, highlighting its robustness across cohorts, which is a key factor for deployment in clinical settings.

\begin{table}[]
\centering
\caption{Comparison results by using CC data as external validation on corresponding scenario models trained with MIMIC data for Random Survival Forest (RSF), traditional Competing-risk Regression (CR), and our proposed FCRN. CSM: Cause-specific model; SDM: Sub-distribution model. Cause 1: ICU readmission; Cause 2: Death.}
\label{tab:ccf_val}
\resizebox{0.8\textwidth}{!}{%
\begin{tabular}{ccccccccc}
\toprule\hline
 & \multicolumn{8}{c}{IBS (CC external dataset)} \\ \cline{2-9} 
 & \multicolumn{4}{c|}{Half year prediction} & \multicolumn{4}{c}{One year prediction} \\ \cline{2-9} 
 & \multicolumn{2}{c}{Cause 1} & \multicolumn{2}{c|}{Cause 2} & \multicolumn{2}{c}{Cause 1} & \multicolumn{2}{c}{Cause 2} \\ \cline{2-9} 
 & CSM & SDM & CSM & \multicolumn{1}{c|}{SDM} & CSM & SDM & CSM & SDM \\ \hline\hline
\multicolumn{9}{c}{Scenario 1: Complete dataset with three functional variables} \\ \hline
\multicolumn{1}{c|}{RSF} & 0.173 & 0.173 & 0.056 & \multicolumn{1}{c|}{0.056} & 0.209 & 0.209 & 0.069 & 0.069 \\
\multicolumn{1}{c|}{CR} & 0.216 & 0.255  & 0.147 & \multicolumn{1}{c|}{0.167} & 0.243 & 0.300 & 0.169 & 0.198 \\
\multicolumn{1}{c|}{FCRN} & 0.131 & 0.127 & 0.057 & \multicolumn{1}{c|}{0.058} & 0.156 & 0.156 & 0.070 & 0.070 \\ \hline
\multicolumn{9}{c}{Scenario 2: Missing dataset with three functional variables} \\ \hline
\multicolumn{1}{c|}{RSF} & 0.167 & 0.167 & 0.057 & \multicolumn{1}{c|}{0.057} & 0.200 & 0.200 & 0.070 & 0.070 \\
\multicolumn{1}{c|}{CR} & 0.181 & 0.211 & 0.162 & \multicolumn{1}{c|}{0.168} & 0.208 & 0.240 & 0.167 & 0.176 \\
\multicolumn{1}{c|}{FCRN} & 0.123 & 0.124 & 0.058 & \multicolumn{1}{c|}{0.058} & 0.147 & 0.153 & 0.070 & 0.070 \\ \hline\bottomrule
\end{tabular}%
}
\end{table}

These findings highlight FCRN's clinical potential in ICU settings. By integrating functional vital signs with static predictors and imputing missing data end-to-end, FCRN enables precise risk prediction for readmission and mortality. For instance, lower IBS for Cause 1 (ICU readmission) suggests FCRN could guide discharge planning, identifying high-risk patients for targeted interventions like enhanced monitoring or follow-up, potentially reducing readmission rates (0.9\%–19\% as reported) and associated costs. In Cause 2 (mortality), improved predictions support early palliative care or resource allocation. External validation on CC data confirms applicability across health systems, addressing EHR heterogeneity and promoting equitable critical care. Future clinical trials could integrate FCRN into decision-support tools to quantify impacts on patient outcomes and hospital efficiency.

\section{Discussion and conclusions}

In this paper, we propose the FCRN, a deep learning framework for competing risk modeling in diverse ICU datasets. By incorporating a Basis Layer and gradient-based missing value imputation into a unified architecture, FCRN extends discrete-time survival analysis to more closely reflect real-world clinical conditions. The framework supports both cause-specific and sub-distribution hazard estimation, enabling flexible and interpretable risk modeling for multiple competing events.

We validated FCRN on simulated datasets and two large real-world ICU datasets. Across experiments, FCRN consistently outperformed traditional models and tree-based methods in terms of the Integrated Brier Score, particularly in settings with complex functional inputs and higher levels of missingness. Notably, FCRN demonstrated strong generalization on external validation using the Cleveland Clinic dataset, supporting its potential for cross-cohort deployment.

These findings highlight FCRN’s clinical potential in ICU settings. By integrating functional vital signs with static predictors and imputing missing data end-to-end, FCRN enables more precise risk prediction for readmission and mortality. Improved prediction of readmission risk can inform discharge planning and targeted post-discharge monitoring, while mortality risk prediction may support timely palliative care referral and resource allocation.

This work has limitations. First, although we considered both cause-specific and sub-distribution hazards, the choice between these two paradigms and related hyperparameter tuning remains task- and data-dependent. Second, the Basis Layer currently assumes sufficient temporal sampling density; extreme sparsity or misalignment across sensors may require additional pre-processing or model adaptations. Third, while our imputation strategy is integrated and gradient-based, its performance depends on assumptions encoded in the learning dynamics.

Overall, FCRN provides a practical, data-integrated approach to competing risks survival analysis in critical care, advancing prognostic modeling where functional signals, static predictors, and missing data coexist. Future research will explore: (i) uncertainty quantification for both predictions and imputations; (ii) incorporation of additional modalities (e.g., clinical notes and imaging) using multi-modal encoders; and (iii) domain adaptation to further improve cross-hospital transportability.

\section*{Acknowledgments}
The research is supported in part by NIH 1R01GM152717-01.

\newpage
\bibliographystyle{elsarticle-num-names} 
\bibliography{ref}





\end{document}